\documentclass{article} 
\usepackage{iclr2027_conference,times}

\usepackage{amsmath,amsfonts,bm}

\def\eqref#1{equation~\ref{#1}}

\def\1{\bm{1}}

\DeclareMathAlphabet{\mathsfit}{\encodingdefault}{\sfdefault}{m}{sl}
\SetMathAlphabet{\mathsfit}{bold}{\encodingdefault}{\sfdefault}{bx}{n}

\usepackage{hyperref}
\usepackage{url}
\usepackage{booktabs} 
\usepackage{multirow} 
\usepackage{graphicx}
\usepackage{enumitem}

\title{
Online Dense Reinforcement Learning with Planning-Grounding Agent for UI Navigation on Live IOS Apps
}

\author{Peitong Duan\thanks{First author} \\
Apple\\
San Francisco, CA, USA \\
\texttt{p\_duan@apple.com} \\
\And
  Khashayar Gatmiry\footnotemark[1] \\
Apple \\
Berkeley, CA, USA \\
\texttt{s\_gatmiry@apple.com} \\
\And
 Jeffrey Nichols \\
Apple \\
Seattle, CA, USA \\
\texttt{jwnichols@apple.com} \\
\And
Samy Bengio \\
Apple \\
Cupertino, CA, USA \\
\texttt{bengio@apple.com} \\
\And
 Emmanuel Abbe \\
Apple \\
Zurich, Switzerland \\
\texttt{e\_abbe@apple.com} \\
}

\iclrfinalcopy 
\begin{document}

\maketitle

\begin{abstract}
While frontier vision-language models demonstrate strong capabilities in user interface (UI) navigation, their high inference costs prohibit everyday use. Conversely, smaller models trained via supervised fine-tuning (SFT) or offline reinforcement learning often memorize static trajectories, leading to brittle execution and an inability to recover from errors. To address these limitations, we propose a novel online reinforcement learning framework that trains UI navigation agents directly on live iOS applications. Our framework decouples the agent into two specialized components: a Planning agent and a UI Grounding model. We rely on an LLM-as-a-Judge to provide comprehensive reward signals, combining dense, step-level feedback to evaluate intermediate progress with a terminal reward for overall task completion. To effectively leverage this dual reward structure, we optimize the agent using Proximal Policy Optimization (PPO) with step-level advantage estimation. Furthermore, we introduce GiGPPO, a novel neighbor-conditioned value estimation method that improves critic stability by leveraging reward signals from visually similar states across concurrent trajectories. To support this online training paradigm at scale, we introduce an efficient infrastructure that orchestrates hundreds of concurrent virtual iOS devices. Evaluations on rigorous benchmarks, including UINavBench, iOSWorld, and AndroidWorld, demonstrate that our online RL trained agents substantially outperform both SFT and terminal-only RL baselines while accelerating training convergence. Notably, our 8B RL agents achieve nearly a sevenfold improvement over the SFT baseline in out-of-distribution generalization to Android environments, while outperforming Gemini-3.5-Flash and tying with Gemini-3.1-Pro on real iOS applications.
\end{abstract}

\section{Introduction}
UI Navigation agents, which are AI systems designed to perceive graphical user interfaces and autonomously execute sequences of actions to achieve user-specified goals, are gaining traction in both industry and academia. Their applications range from automatically completing tasks for users \citep{DBLP:journals/corr/abs-2501-12326, gu2025uivenustechnicalreportbuilding} to A/B testing \citep{10.1145/3772363.3799039}. While frontier models can be quite proficient at UI Navigation \citep{jang2026iosworldbenchmarkpersonallyintelligent, Agrawal_2025_ICCV}, they are expensive, often costing several dollars per task \citep{easyclaw2026manus}, making them impractically expensive for frequent, everyday use. As a result, there has been a surge in interest for training small models for proficient UI navigation \citep{yang2025ferretuilitelessonsbuilding, chai2026uikobeknowledgeorientedbehaviorexploration}. However, current training methods suffer from significant weaknesses. Supervised Fine-tuning (SFT) tends to memorize training trajectories rather than learn generalizable strategies, resulting in brittle execution that struggles to recover from errors \citep{PAE, NEURIPS2024_1704ddd0, qi2024webrl, yangui}. Existing Reinforcement Learning (RL) approaches largely rely on single-step RL that mirrors these same SFT limitations \citep{yang2025ferretuilitelessonsbuilding, yangui} or rely on only a terminal reward at the end of a trajectory, providing a weak, sparse signal \citep{NEURIPS2024_1704ddd0, wang2025distrl, qi2024webrl, zhou2025mai}.

To address these challenges, we propose a novel framework, trained at a large scale on real iOS applications, that decouples the agent into two specialized components: a Planning Agent and a UI Grounding Model. Rather than training a single model to jointly handle planning and spatial localization, our architecture allows each component to be optimized for its specific objective. The Planning Agent observes the screen, maintains a history of past actions, and utilizes Chain-of-Thought reasoning to produce both an immediate step and a long-term plan. The immediate step is then passed to the Grounding Model, which translates it into a precise action and corresponding coordinates for on-device execution. Additionally, the Planning Agent's action space includes an \textsc{Ask\_User} primitive to handle underspecified tasks by requesting user preference specifications. Because grounding is a largely deterministic mapping problem effectively solved via SFT, this decoupling allows us to reserve computationally expensive RL for the Planning Agent, where open-ended reasoning and adaptation provide the most benefit.

We optimize the Planning Agent in an online RL setting, centered on a robust LLM-as-a-Judge reward mechanism. To overcome the sparsity of terminal-only feedback, we combine two complementary signals. A Dense Reward evaluates whether each individual step makes meaningful progress while penalizing hallucinations and unproductive loops. A Terminal Reward assesses overall task completion. This dense, step-level feedback provides a rich signal throughout the trajectory, giving the model useful feedback even when terminal success is rare under the current policy. It further encourages exploration and explicitly teaches the agent to recover from unexpected states rather than failing when it deviates from a golden path. To effectively leverage this dense reward structure, we optimize the agent using Proximal Policy Optimization (PPO) with step-level Generalized Advantage Estimation (GAE). To prevent the agent from artificially prolonging trajectories to exploit step-level rewards, we use an asymmetric discounting strategy in our reward function that balances local process supervision with discounted dense rewards, combined with a terminal reward for successful task completion. Furthermore, we introduce GiGPPO, a novel neighbor-conditioned value estimation method that improves critic stability by incorporating reward signals from visually similar states across concurrent trajectories.

Because online RL requires continuous interaction with real UI environments, we built a highly scalable device management system to support this training paradigm. By orchestrating a large pool of cloud-hosted iOS devices and utilizing rapid virtual machine snapshots, our infrastructure enables hundreds of parallel live trajectories. This allows the agent to generate fresh rollouts across concurrent environment instances without bottlenecking on sequential device interactions. To our knowledge, this work represents the first time UI navigation agents have been trained at-scale entirely on live iOS devices interacting with real iOS applications.

We evaluated our trained agent system on public benchmarks, including UINavBench and iOSWorld, as well as AndroidWorld to test out-of-distribution performance. Our experiments demonstrate that our 8B online RL agents substantially improve upon the SFT baseline. Most notably, the RL agents achieve nearly a sevenfold higher success rate than the SFT baseline on AndroidWorld, proving that our framework successfully teaches generalizable navigation strategies rather than simply memorizing iOS-specific trajectories. Furthermore, both our dense reward formulation and GiGPPO accelerate training convergence, and on UINavBench, a rigorous benchmark of real iOS applications, our RL agents outperform Qwen3.8-27B and Gemini-3.5-Flash while tying with Gemini-3.1-Pro.
In summary, our key contributions are as follows:
\begin{itemize}[itemsep=0pt]
    \item \textbf{Online RL planning-grounding framework on Live iOS:} We propose a decoupled planning-and-grounding architecture trained at scale entirely on live iOS devices, utilizing LLM-as-a-Judge to provide both dense and terminal rewards.
    \item \textbf{Asymmetric Reward Discounting and GiGPPO:} We identify and resolve reward hacking in step-level RL by introducing an asymmetric discounting strategy that balances local process supervision with undiscounted terminal task completion. We also introduce a novel neighbor-conditioned value estimation method (GiGPPO) that stabilizes the PPO critic by incorporating reward signals from visually similar states across concurrent trajectories.
    \item \textbf{Empirical Generalization:} We demonstrate that our 8B RL agents substantially outperform SFT and terminal-only RL baselines, generalize more effectively to out-of-distribution Android environments, and outperform Qwen3.8-27B and Gemini-3.5-Flash, while tying with Gemini-3.1-Pro on a benchmark of real iOS applications.
\end{itemize}

\section{Related Work}
\subsection{UI Navigation Agents}
The rise of vision-language models (VLMs) has produced a wave of research on UI navigation agents. Early agents simply prompted a frontier VLM with no training \citep{zhang2025litewebagent}; SeeAct-V \citep{zheng2024seeact} improved on this by pairing a zero-shot frontier planner with a separately trained visual grounding model, but left the planner itself untouched. Running a frontier model on every step is expensive and slow, which has pushed the field toward smaller VLMs that may fit on-device \citep{yang2025ferretuilitelessonsbuilding}, but a small model cannot rely on a frontier model's zero-shot competence, so it has to be trained. One training approach is supervised fine-tuning (SFT) on human or model-collected demonstrations \citep{yang2025ferretuilitelessonsbuilding, chai2026uikobeknowledgeorientedbehaviorexploration}. SFT training is mostly on successful trajectories, so it learns to imitate the next action along a path that is already going right rather than a generalizable strategy; it rarely sees what an error looks like or what to do once one occurs, and tends to memorize training trajectories rather than adapt to states outside them, resulting in poor error recovery and weak generalization to out-of-distribution apps and tasks \citep{PAE, Agrawal_2025_ICCV, qi2024webrl, yangui}.

To address the limitations of SFT, researchers have started exploring RL methods. However, many of these approaches mirror SFT's limitations through single-step RL that rewards similarity to golden traces \citep{yang2025ferretuilitelessonsbuilding, yangui}, or by relying on offline human corrections. For example, UI-TARS \citep{qin2025uitarspioneeringautomatedgui} fine-tunes on human-relabeled ``reflection'' pairs via DPO rather than learning recovery through live interaction. For online training methods, several systems use an independent judge (such as a VLM) to evaluate trajectories without needing reference traces \citep{NEURIPS2024_1704ddd0, wang2025distrl, qi2024webrl, zhou2025mai}. However, these approaches provide a terminal reward at the end of the trajectory, which provides a sparse training signal and does not give the agent credit for intermediate steps that make real progress toward a goal it ultimately fails to finish. While some of these training frameworks add dense signals like hand-coded repetition penalties \citep{zhou2025mai}, they still lack a mechanism to reward partial success. Similarly, Qwen-UI-Agent \citep{zhou2026qwen_ui_agent} trains an end-to-end agent across approximately 10,000 concurrent environments (mobile (Android), web, and desktop). Its multistage pipeline includes offline RL training targeting per-step action accuracy with a set of hand-curated action failure types. However, this per-step accuracy focuses on localized checks (e.g., grounding accuracy and action repetition) rather than assessing genuine progress toward the goal. This offline phase is followed by online training using only terminal rewards. In contrast, our framework provides both dense progress evaluation and terminal task-completion rewards within a single online RL stage, using an LLM-as-a-Judge without requiring manually curated error taxonomies or pattern-specific training data. Across this line of work, group-relative methods have become the default: GRPO or a close variant is used by \citet{yang2025ferretuilitelessonsbuilding}, \citet{yangui}, \citet{zhou2025mai}, \citet{zhou2026qwen_ui_agent}, and \citet{tang2026clawgui}. A consequence is that no prior GUI-agent system learns a value function for step-level credit assignment; dense signals, where present, are aggregated into trajectory-level returns or into subgroup averages over repeated states. Our framework instead trains a critic that consumes the dense channel directly, which is what makes the asymmetric discounting of Section~\ref{sec:rl-optimization} possible.

Concurrent with our work, ClawGUI-RL \citep{tang2026clawgui} combines a terminal outcome reward with a dense step-level reward from a Process Reward Model. It trains a single end-to-end policy for both planning and grounding using GiGPO on Android emulators at a scale of 64 concurrent environments. We differ in three key respects. First, we decouple a specialized RL-trained planner from a frozen SFT grounder rather than training end-to-end. Second, while GiGPO forms a step-level baseline by averaging returns of exactly recurring states, which degenerate into singletons in our setting, we route this cross-trajectory signal through a learned value function (Section~\ref{sec:rl-optimization}). Third, we train entirely on live iOS devices at a scale of hundreds of concurrent trajectories, representing the first at-scale RL training of UI agents on real iOS applications.

\subsection{UI Navigation Benchmarks}
Early UI agent benchmarks evaluated models offline on static screenshot–action pairs \citep{3666122.3667342, li-etal-2020-mapping}, which cannot assess sequential decision-making or error recovery. The field has since shifted toward live-environment evaluation: MiniWob++ \citep{zheran2018reinforcement} and WebArena \citep{zhou2023webarena} require agents to act in real browser sessions, OSWorld \citep{xie2024osworld} evaluates desktop tasks across multiple operating systems, and AndroidWorld \citep{rawles2024androidworld} and MobileWorld \citep{kong-etal-2026-mobileworld} target Android apps with programmatic state verification to measure task success. iOS evaluation has been comparatively underserved until recently, with UINavBench \citep{Agrawal_2025_ICCV} and iOSWorld \citep{jang2026iosworldbenchmarkpersonallyintelligent} introducing live-device evaluation with VLM-based success judges. While most of these benchmarks focus on single-app tasks, AndroidWorld, UINavBench, and iOSWorld include workflows that require planning across multiple applications.
\section{UI Navigation Agent System and Training}
\begin{figure}
    \centering
    \includegraphics[width=\linewidth]{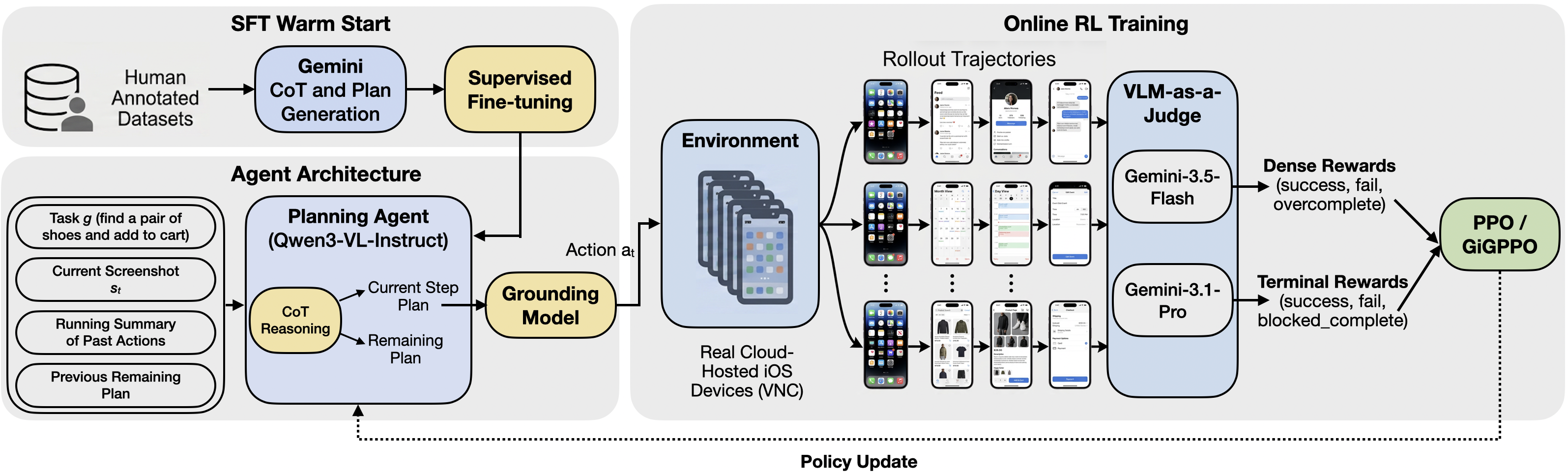}
    \caption{Overview of the iOS agent architecture and training setup.}
    \label{fig:systemdiagram}
\end{figure}
\subsection{Problem Formulation}
We treat UI navigation as a sequential decision process. Given a natural-language task instruction $g$ (e.g., ``Search for a pair of shoes and add it to cart''), an agent repeatedly observes the current screen and executes a single action, continuing until the task is complete. At each step $t$, the agent observes the current state, which consists of the current screenshot $s_t$ and a running summary of its past actions. The agent then outputs one action $a_t$ from a fixed set of primitives: \textsc{Tap}, \textsc{Swipe}, \textsc{Text\_Entry}, \textsc{Navigate\_Home}, \textsc{Long\_Press}, \textsc{Ask\_User}, and \textsc{Task\_Complete}. 

The \textsc{Ask\_User} action allows the agent to request preference-related clarification for underspecified parts of a task. In the shoe search example, the agent might ask the user what type of shoes to search for and which specific option from the search results to add to the cart. Executing $a_t$ transitions the environment to a new state, in a way that depends on the app that the agent does not control directly. Because the instruction states only the final goal rather than the steps to reach it, the agent must determine its own sequence of screen-specific actions and dynamically revise this sequence whenever the environment deviates from expectations. An episode ends when the agent selects \textsc{Task\_Complete} or reaches the maximum step limit.
\subsection{Architecture}\label{sec:architecture}
To produce $a_t$, the agent must determine what to do on the current screen (planning) and identify the exact x,y-coordinates to execute the planned action (grounding). Unlike prior work that combines both functions into a single model, we decouple the agent into two separate models for planning and grounding. Figure \ref{fig:systemdiagram} illustrates the agent architecture as well as the training setup.

At each step, the Planning Agent outputs a plan for the current screenshot, preceded by chain-of-thought (CoT) reasoning \citep{3600270.3602070}. The agent also outputs a remaining plan (a set of high-level goals required to finish the task), which is similarly preceded by its own CoT reasoning. The inputs to the Planning Agent include the task instruction, the current screenshot, a history of previous step plans with their reasoning, and the remaining plan from the prior step.

The current step plan is then passed to the Grounding Model, which maps it to one of the action primitives and the specific x,y-coordinates on the screen to execute on device. We decouple these components because planning and grounding represent fundamentally different challenges, and splitting them maximizes model performance in both aspects. Grounding requires precise spatial understanding to map a specific intent to UI coordinates, a task that can be trained highly effectively with SFT \citep{gou2024uground, lin2025showui}. Planning, conversely, is an open-ended task that requires understanding the overall goal, tracking completed actions, adapting to dynamic changes in the UI, and recovering from errors when unexpected states arise. This open-ended reasoning and need for continuous adaptation make RL a highly suitable training method for the Planning Agent. By separating the two, we can reserve computationally expensive RL for the planning phase where it provides the most benefit. Appendix \ref{sec:grounding} compares the performance of the Planning Agent when paired with different Grounding Models.

\textbf{Virtual device environment.} The agent operates on cloud-hosted iOS virtual machine devices that are provisioned through cloud-based service, which also handles device lifecycle management and automatic retirement of degraded sessions. Devices are driven over a VNC session via a typed action interface (e.g., tap, swipe, type, long press, home, wait) that mirrors physical-device behavior. Each device is configured once with the necessary apps installed and signed in with an Apple ID. This verified ready state is then captured as a virtual machine snapshot, which the device can be reverted to at any time through the cloud-based service. Episodes end by rolling back to the snapshot rather than tearing the session down, which reduces restart latency to roughly 10 seconds. This combination of on-device fidelity and rapid snapshot-based reuse is necessary for our online RL training with high throughput. During training, our system is able to run 512 trajectories simultaneously, each with its own independent device and agent-environment loop.

\subsection{SFT Warm Start}
Before RL, we warm-start the Planning Agent from Qwen3-VL-8B-Instruct \citep{Qwen3-VL} using supervised fine-tuning on human-annotated task trajectories. The training data are drawn from three separate datasets. Two are collected in-house: one covering first-party iOS apps and another covering third-party iOS apps. Each pairs a task goal with a human-executed sequence of screens and the corresponding actions taken. The third is the public GUIOdyssey dataset \citep{lu2025guiodyssey}. Although it consists of Android apps, we include it specifically for its multi-app trajectories (tasks requiring navigation across several apps). This addresses a gap in our iOS datasets, which only contain single-app tasks and lack app-switching behavior.

None of the three datasets natively record reasoning. They provide only the goal, the sequence of screens, and the current plan at each step, lacking both CoT reasoning and long-term plans. To produce the reasoning and remaining-plan targets that the Planning Agent is designed to output (Section~\ref{sec:architecture}), we use Gemini \citep{Gemini3Pro2025} to generate the chain-of-thought reasoning consistent with each step plan, as well as the plan for the remaining steps and its corresponding reasoning.

\subsection{Online Reinforcement Learning Setup}\label{sec:rl}
To overcome the limitations of offline training on static datasets seen in prior work, we optimize the Planning Agent using online RL training. This approach employs an LLM-as-a-Judge as a fast and robust method to generate real-time rewards that adapt to the dynamic UI environment. During training, the agent continuously interacts with live virtual iOS devices to generate fresh rollouts. This active interaction allows the model to experience the direct consequences of its actions, navigate unexpected screen states, and learn robust error recovery strategies using online reinforcement learning algorithms, specifically Proximal Policy Optimization (PPO) and our novel neighbor-conditioned variant, GiGPPO. Rewards are computed at both the step and trajectory levels, and all trajectories collected at a step are batched into a single synchronous on-policy update. For \textsc{Ask\_User} steps that require user input, we use Gemini-3.5-Flash to provide realistic user responses. We use this simulated user during both RL training and the benchmark evaluations in Section \ref{sec:experiments}.

\textbf{Training Data}
For online RL training, we randomly sampled tasks and apps from the existing iOS datasets used during SFT. Because the multi-app tasks in GUIOdyssey are on Android, they could not be directly executed in our iOS environment. To address this, we synthetically generated multi-app iOS tasks using a two-stage pipeline to ensure diversity. First, we prompted Gemini to identify logical pairs of iOS apps for cross-app workflows. For example, pairing Weather and Calendar enables tasks like finding a day with clear weather to schedule an outdoor event. Second, for each identified app pair, we generated a diverse set of multi-app tasks, using single-app examples from our iOS datasets as in-context references to guide the generation process. To ensure realism and feasibility of the generated tasks, one author manually verified each one, making corrections as needed.

\textbf{Reward Design}
Studies have shown that LLM-as-a-judge is an effective way to provide RL rewards during training \citep{zhou2025mai, qi2024webrl, wang2025distrl, PAE}. We use this approach as an efficient and robust way to supply live rewards during online training. To ensure sufficient signal, we provide both dense and terminal rewards, and design each reward judge to handle a variety of scenarios. Because our LLM-as-a-judge reward requires no ground-truth action sequence, it generalizes to any app/task pair rather than only those with human-annotated golden traces and enables RL training directly on the synthetic multi-app tasks.

Since the dense reward judge is called at every step, we use Gemini-3.5-Flash. Given the task, action history (with the corresponding dense judge results), current plan, reasoning, and before/after screenshots, the judge returns one of three outcomes: success, fail, or overcomplete. To achieve a \emph{success}, the before/after screenshots must show progress towards completing the overall task, and agent's reasoning and action must both align with the visual state of the UI; accidental progress from hallucinated reasoning is penalized as a failure. The judge also penalizes unproductive loops by checking for repeated failed actions. The \emph{overcomplete} label penalizes the agent for continuing to act after the task is already satisfied, teaching the policy to recognize task completion. Finally, a separate dense judge evaluates \textsc{Ask\_User} actions, using only the current screenshot to verify whether the requested information is genuinely underdetermined at that step.

Similar to \citet{PAE}, our terminal judge evaluates task success using the trajectory's final three screenshots. However, we additionally provide the full action history, along with prior dense-judge verdicts. The judge outputs one of three outcomes: \emph{success}, \emph{fail}, or \emph{blocked\_complete}. The \emph{blocked\_complete} label rewards trajectories that correctly execute all steps up to an unavoidable external barrier, such as a paywall or CAPTCHA, preventing the agent from being penalized for unactionable states. We use Gemini-3.1-Pro for the terminal judge.

\subsection{Online Reinforcement Learning Optimization}\label{sec:rl-optimization}
While GRPO is effective for trajectory-level optimization and is commonly used to train UI Navigation Agents \citep{yangui, zhou2025mai, zhou2026qwen_ui_agent}, it is less suited to exploit rich intermediate supervision: dense rewards must be aggregated with the terminal reward before computing a group-relative advantage, obscuring their temporal structure. Alternatively, one can estimate a separate GRPO advantage at each state by sampling multiple candidate actions, as in Verifiable Process Rewards (VPR) \citep{yuan2026verifiable}, but this requires multiple additional samples per state and does not naturally incorporate the eventual task outcome into the local advantage. We instead use Proximal Policy Optimization (PPO), whose critic can directly exploit dense rewards. Recently, StepSearch \citep{zheng-etal-2025-stepsearch} combined step-wise and terminal rewards using PPO with Generalized Advantage Estimation (GAE) and an undiscounted return ($\gamma=1$). In our setting, however, this requires the critic at each state to predict the expected cumulative contribution of potentially distant future dense rewards which is a difficult task for the critic. 

\textbf{Reward Discounting}
Furthermore, as observed in the run detailed in Appendix~\ref{sec:rl-setup}, an undiscounted return can lead to reward hacking. With a discount factor of $\gamma=1$, every future dense reward is valued equally to an immediate one, removing the agent's incentive to complete the task efficiently. Because dense credit accrues at each step up to the maximum step budget, the undiscounted return is maximized simply by taking more steps that appear locally correct. Consequently, the agent learns to artificially prolong trajectories to collect positive dense rewards at each step rather than completing the task and terminating.
We instead use $\gamma=1/2$ for the dense component, making its value target predominantly local and easier to estimate. Therefore, denoting the terminal reward by $R_T$ and the dense reward at intermediate step $t$ for $1\leq t\leq T$ and a given state $s_{t-1}$ of actions already taken by the agent, the critic model tries to fit the following value function at time step $t$:
\begin{align*}
V^\pi(s_t)=
\mathbb{E}_\pi\left[
R_T+\sum_{k=t}^{T}\gamma^{k-t}r_k^{\mathrm{dense}}
\mid s_t
\right], \ 
\gamma=\tfrac12.
\end{align*}
For the advantage we use the GAE estimator with $\lambda=1$, which collapses it to a Monte Carlo estimator: the advantage at step $t$ is the return actually realized by the rolled-out trajectory, $R_T+\sum_{k=t}^{T}\gamma^{k-t}r_k^{\mathrm{dense}}$, minus the critic's estimate $V^\pi(s_t)$ of the value at that state. The advantage is thus unbiased in the critic, trading the bias that a bootstrapped estimate would introduce for the variance of the sampled return.
Importantly, we use a linear model on top of the critic's hidden embedding on the last token before the tokens of the action $a_{t+}$ of time step $t+1$. We avoid fitting the average embedding of on top of all the action tokens for $a_{t+1}$ as this violates the definition of value function at state $t$ which is not supposed to be conditioned on any information about $a_{t+1}$.

Discounting $R_T$ as well would reduce its contribution to an early action by $\gamma^{T-t}$, making the task-level signal negligible on long trajectories. Preserving it at a scale comparable to the dense component is particularly important when the current policy deviates from the locally preferred action of the dense-reward judge but follows an alternative, potentially longer, path that can still complete the task. The undiscounted terminal reward preserves credit for such successful strategies, while dense rewards provide richer local guidance when terminal success is sparse. Our asymmetric return thus combines local process supervision with an optimization signal that remains anchored to the actual evaluation objective: successful task completion.


\textbf{Neighbor-conditioned value estimation (GiGPPO)} GRPO estimates the value of the states along a trajectory by comparison with the other trajectories sampled for the same task, but the comparison is trajectory-level: returns are aggregated before the advantage is formed, so dense rewards collapse into a single scalar and their temporal structure is lost. GiGPO \citep{feng2025gigpo} refines this idea by exploiting repeated states, collecting states that recur across the trajectories of a task into subgroups and estimating the value of a state as the average reward within its subgroup, which yields a step-level baseline without a critic. In our setting, however, an identical screen almost never recurs: with $G=4$ or $G=8$ rollouts per task, subgroups defined by exact state equality are essentially singletons, while relaxing equality to screen proximity would average over states that are merely similar and pass that bias directly into the critic's value estimates. 

To address this, alongside our PPO implementation, we introduce and evaluate an alternative method: \textit{GiGPPO}, which retains the underlying intuition that the reward observed at a similar state in a sibling trajectory is informative about the value of the current state, but route it through the critic rather than through averaging. Each state is paired with the state whose $64$-bit perceptual hash lies closest in Hamming distance among the sibling trajectories of the same task; we call this state its \emph{neighbor}. A small multi-layer perceptron then predicts $V^\pi(s_t)$ from the hidden embeddings of the state and of its neighbor, the neighbor's return (its discounted dense sum plus undiscounted terminal reward), and the Hamming distance between the two screens, the last of which lets the critic calibrate how far the neighbor's reward should be trusted: at distance zero the two states are the same screen and that reward is nearly a direct observation of the quantity being predicted, whereas at large distance the screens differ and its contribution should be discounted. The state's own reward is never an input, so the regression target cannot leak. Conditioning on the neighbor in this way gives the critic strictly more information than standard PPO, where $V^\pi(s_t)$ must be inferred from $s_t$ alone.

\textbf{Reward structure} For the PPO and GiGPPO runs, we utilize a dual-channel reward structure. The \emph{dense} channel scores each step and is discounted at $\gamma_{\text{dense}}=0.5$: $+1.0$ for a judged \emph{success}, $0.0$ for a judged \emph{fail} or an ungroundable plan, $-1.0$ for an empty plan or a generation truncated at the length limit, and $-2.0$ for acting after the task is already complete (i.e., \emph{overcomplete}). The \emph{terminal} channel is paid once and carried undiscounted to every step, gated on what the agent declared: $+2.0$ for completing the task and declaring \emph{Task\_Complete}, $+0.5$ for encountering \emph{blocked\_complete}, $-0.75$ for exhausting the step budget, $-1.0$ for a false completion claim, and $-2.0$ for ending on unparseable or empty output. Since the dense return is bounded by $1/(1-\gamma_{\text{dense}})=2.0$, a trajectory that succeeds at every step earns exactly one terminal success worth of dense credit, so neither channel dominates. Ordinary failure is left unpunished to keep exploration cheap, but unusable output is charged at terminal scale, since a single discounted dense charge cannot outweigh a guaranteed terminal cell of zero and an unusable action scored at zero is gradient-equivalent to a genuine one while being cheaper to emit.

Trajectories lost to infrastructure rather than behavior, such as an application that never renders, a failed app installation, a dropped device connection, or an exceeded generation deadline, are masked instead of penalized: they enter neither the policy gradient nor the critic regression, cannot serve as neighbors, and are excluded from the statistics used to whiten advantages, which matters because their zero reward values would deflate the standard deviation and inflate every genuine advantage. Remaining optimization and hyperparameter details are given in Appendix~\ref{sec:ppo-setup}.

\textbf{Critic Training} For both the standard PPO and GiGPPO runs, the value model (i.e., critic) is initialized from the supervised fine-tuning checkpoint with a fresh scalar value head. We found that training the critic and the policy jointly from the start is highly unstable. Because the initial value estimates are uninformative, the resulting advantages are dominated by value error rather than actual reward signals, causing the policy to take large, disruptive early updates. To mitigate this, we warm start the critic for 15 steps. During this initial phase, the policy is held frozen while rollouts are collected, and the value function is regressed directly onto the observed returns. Only after this warm-up period do we begin joint training. Furthermore, the critic uses a learning rate of $1\times10^{-4}$, which is two orders of magnitude larger than the policy's learning rate of $1\times10^{-6}$. This discrepancy is intentional. The value function must rapidly track a moving target and is ultimately discarded at the end of training, so it can afford to update quickly. In contrast, the policy is the final artifact being preserved and must be updated deliberately slowly to avoid destroying its supervised fine-tuning initialization.

\section{Experiments}\label{sec:experiments}
\subsection{Benchmarks}
Because our training focused primarily on iOS apps, we evaluated our agents on two public iOS benchmarks: UINavBench \citep{Agrawal_2025_ICCV} and iOSWorld \cite{jang2026iosworldbenchmarkpersonallyintelligent}. UINavBench consists of 116 tasks of varying difficulty across 36 real iOS apps staged with user data. iOSWorld comprises 133 tasks across 26 synthetic iOS apps, all staged with data reflecting a single user persona. Both benchmarks include a mix of single-app and multi-app tasks. Additionally, iOSWorld features 46 tasks specifically designed to test the agent's personalization capabilities, requiring the agent to discover and act upon the user persona's underlying preferences. For both benchmarks, task success is evaluated using a Vision-Language Model (VLM) that verifies whether a specific set of requirements is met by the agent's output trajectory. To assess the generalizability of our iOS-trained agent, we also evaluate it on out-of-distribution Android apps using the AndroidWorld benchmark \citep{rawles2024androidworld}, which consists of 116 tasks across 20 real applications.
\subsection{UI Grounding Models}
For grounding, we utilized an internal 3B dense pretrained model finetuned on a large human-annotated dataset that maps step-level instructions to action primitives and precise x,y-coordinates on real iOS screenshots. This model achieves over 90\% accuracy on its held-out test set. However, because it supports only directional swipes (e.g., ``swipe up'') rather than precise location-based swiping, we route all swipe instructions to Gemini-3.5-Flash, which has demonstrated strong UI grounding capabilities \citep{jang2026iosworldbenchmarkpersonallyintelligent} and supports location-based swiping.  We ran the RL training using this grounding setup, but experimented with various grounding configurations on the benchmarks in Appendix \ref{sec:grounding}. We report the task success rates with the optimal grounding configuration for each benchmark in Table \ref{tab:benchmark-results}.


\subsection{Results}\label{sec:results}
\begin{table}[h]
\centering
\begin{tabular*}{\textwidth}{@{\extracolsep{\fill}}lccc}
\toprule
\multirow{1}{*}{} & \textbf{UINavBench} & \textbf{iOSWorld} & \textbf{AndroidWorld} \\
\midrule
SFT Baseline   & 16.7\% & 4.5\% & 4.7\% \\
PPO (terminal only), Step 75   & 17.6\% & 3.0\% & 9.6\% \\
PPO (terminal only), Step 168   & 17.6\% & 5.3\% & 6.9\% \\
PPO (terminal only), Step 336   & 21.6\% & 7.5\% & 19.8\% \\
PPO, Step 70   & 14.7\% & 6.0\% & 11.2\% \\
PPO, Step 168 & 25.5\% & 6.0\% & 20.7\% \\
PPO, Step 336 & \textbf{32.4\%} & 2.3\% & \textbf{32.8\%} \\
GiGPPO, Step 75   & 21.6\% & 5.3\% & 7.8\% \\
GiGPPO, Step 168   & 30.4\% & 4.5\% & 18.1\% \\
GiGPPO, Step 336   & 28.4\% & \textbf{9.8\%} & 31.9\% \\
\midrule
UI-TARS-72B  & 11.0\% & N/A & 46.6\% \\
Qwen3.8-27B      & 19.6\% & 19.5\% & 49.1\% \\
Gemini-3.5-Flash  & 28.4\% & 12.9\% & 57.8\% \\
Gemini-3.1-Pro    & 32.4\% & 39.5\% & 60.8\% \\
\bottomrule
\end{tabular*}
\caption{Success rates across UINavBench, iOSWorld, and AndroidWorld. We compare our SFT baseline against our RL-trained agents (PPO and GiGPPO), including an ablation of PPO trained with only terminal rewards. For each RL checkpoint, we report the number of update steps, where Steps 168 and 336 correspond to one and two epochs, respectively. We also include results for the UI-TARS agent and three frontier models (Qwen3.8-27B, Gemini-3.5-Flash, and Gemini-3.1-Pro) evaluated using our agent setup. UI-TARS was not evaluated on iOSWorld.}
\label{tab:benchmark-results}
\end{table}
Table \ref{tab:benchmark-results} presents the performance of our SFT baseline alongside the PPO and GiGPPO checkpoints across different training durations. To measure the impact of dense rewards, we also include results for PPO trained with only terminal rewards. Each RL checkpoint in the table is labeled with the number of update steps. We compare our approach against the end-to-end UI-TARS model, as well as three frontier models (Qwen3.8-27B, Gemini-3.5-Flash, and Gemini-3.1-Pro). We evaluate these frontier models using our decoupled architecture, where they act as planners, receiving the same screenshot and prompt inputs as our agent, and are paired with a separate grounding model. For each benchmark, we use the optimal grounding configuration identified in Appendix \ref{sec:grounding}: the RL training grounding setup for UINavBench, and Gemini-3.5-Flash for both iOSWorld and AndroidWorld.

Overall, online RL training yields significant gains over the SFT baseline, with longer training generally producing stronger benchmark performance. Notably, on UINavBench, PPO Step 336 (2 epochs) ties with Gemini-3.1-Pro (32.4\%), three of the RL checkpoints score at least as high as Gemini-3.5-Flash (28.4\%), and six RL checkpoints outperform Qwen3.8-27B (19.6\%), the latest model in the Qwen family known for its strong computer-use capabilities \citep{qwen38}, despite being over three times larger than our 8B RL agents. In addition, every RL checkpoint outperforms the SFT baseline on AndroidWorld, with PPO Step 336 scoring nearly 7 times higher (32.8\% vs. 4.7\%). This demonstrates that our RL framework successfully teaches generalizable navigation strategies that transfer effectively to the out-of-distribution Android environment, despite the agent being trained exclusively on iOS apps during RL training. Appendix \ref{sec:traces} presents traces from the RL and SFT agents on the benchmark tasks, showcasing the RL agents' ability to recover from errors and handle complex workflows.

To evaluate the contribution of our dense reward formulation, we compare standard PPO against the terminal-only PPO ablation. Standard PPO generally outperforms the terminal-only agent at equivalent training steps; notably, the performance of standard PPO at Step 168 (1 epoch) already matches or exceeds that of terminal-only PPO at Step 336 (2 epochs) across benchmarks, indicating that dense rewards provide a richer signal that accelerates training. Furthermore, on UINavBench, GiGPPO at Step 75 and Step 168 achieves comparable performance to standard PPO at Step 168 and Step 336, respectively. Because UINavBench runs on real iOS apps and most closely matches the RL training environment, this accelerated learning under GiGPPO can likely be attributed to the critic effectively leveraging signals from visually similar UI screens for value estimation. 


The benchmark results show modest gains on iOSWorld, with the exception of GiGPPO Step 336, which achieved more than double the success rate of the SFT baseline (9.8\% vs. 4.5\%). One reason for this is that iOSWorld is built on synthetic applications whose visual layouts and structures deviate from the real iOS apps our agents were trained on. Another factor is the greater complexity of the benchmark's tasks, which include user personalization workflows (e.g., ``find my most frequently ordered restaurant and item, then reorder my usual'') and multi-app tasks spanning more than two apps; both categories account for the majority of the benchmark and are out-of-distribution for our agent's training. Interestingly, in later RL checkpoints, we observed that when the agent searched for a synthetic app by name, it sometimes encountered a naming collision with a real iOS app and attempted to install it from the App Store, or assumed the synthetic app was a web service and searched for it in Safari (since the training set included Safari search tasks). We will investigate ways to further improve performance on this benchmark in future work. Finally, while RL substantially improved performance on AndroidWorld, the frontier models still outperformed our RL checkpoints. However, the frontier models also performed considerably better on AndroidWorld than on iOSWorld. While this may partly reflect differences in task difficulty, it is also likely driven by the widespread availability of large public Android UI datasets \citep{NEURIPS2023_bbbb6308, lu2025guiodyssey, li2024effects}, which were likely included in their pretraining, whereas public iOS datasets remain scarce. Similarly, UI-TARS's strong performance on AndroidWorld is attributed to its training specifically on Android data \citep{DBLP:journals/corr/abs-2501-12326}.


\section{Conclusion}
In this work, we presented a novel online reinforcement learning framework that trains UI navigation agents directly on live iOS applications to overcome the brittleness of supervised fine-tuning and the high inference costs of frontier models. By decoupling the agent into a specialized RL-trained planner and a frozen SFT grounder, we efficiently focused open-ended reasoning capabilities where they are most needed. To provide rich learning signals, we utilized an LLM-as-a-Judge to supply both dense, step-level feedback and terminal task-completion rewards. We optimized this dual reward structure using Proximal Policy Optimization (PPO) with a novel asymmetric discounting strategy to prevent reward hacking, and introduced GiGPPO, a neighbor-conditioned value estimation method that stabilizes the critic using visually similar states across concurrent trajectories. Evaluations across UINavBench, iOSWorld, and AndroidWorld demonstrate that our online RL agents substantially outperform the SFT baseline, while our dense reward formulation and GiGPPO significantly accelerate training convergence compared to terminal-only rewards. Notably, on real iOS applications, our 8B RL agents surpass multiple frontier models and match Gemini-3.1-Pro, while achieving nearly a sevenfold improvement over the SFT baseline in out-of-distribution generalization to Android environments. Ultimately, this work provides a highly effective pathway toward robust, generalizable, and economically viable UI navigation agents.

\section*{AI use statement}
In this work, we used generative AI for the synthetic generation of data, UI grounding, and as an LLM-as-a-Judge for reward functions during RL training. Additionally, we used generative AI tools to write software code, identify relevant literature, and edit the research paper to improve readability. We have reviewed all AI-assisted work. Specifically, all LLM-generated code was verified and tested for correctness by the authors. We take full responsibility for the final content of this work, including the final manuscript and artifacts produced with the aid of generative AI.
\bibliography{iclr2027_conference}
\bibliographystyle{iclr2027_conference}

\appendix
\section{Appendix}

\subsection{PPO Training Details}\label{sec:ppo-setup}
The Planning Agent is initialized from the same 8B vision-language supervised fine-tuning checkpoint used in Appendix~\ref{sec:rl-setup}, and the critic from that checkpoint with a fresh value head. The policy's vision tower is frozen; the critic is adapted with LoRA of rank 32 and $\alpha=64$, excluding the visual modules, beneath the neighbor-conditioned network, whose projection and hidden widths are $128$ and $512$ and which adds about $0.9$M trainable parameters. Each step draws 16 tasks and samples $G=8$ trajectories per task, giving 128 trajectories, with episodes capped at 30 agent steps, prompts at $12288$ tokens and generations at $700$ new tokens per step; rollouts are sampled at temperature 1.0 and evaluation rollouts at temperature 0.7 with 5 samples per task. Advantages use step-level GAE with $\gamma=\lambda=1$ and a dense discount $\gamma_{\mathrm{dense}}=1/2$. The objective is a dual-clipped surrogate with $\epsilon_{\text{low}}=0.18$, $\epsilon_{\text{high}}=0.22$ and $c=3$, together with a KL penalty of $0.01$ to the frozen supervised fine-tuning reference; value updates are clipped at $0.5$. The policy takes two optimizer steps per batch, using a mini-batch of 8 tasks over one epoch, with AdamW at a learning rate of $1\times10^{-6}$, while the critic takes one at $1\times10^{-4}$; both use a weight decay of $0.01$, no warmup, and gradient-norm clipping at $1.0$. The critic is trained alone for the first 15 steps, with the policy frozen, so that advantages are not dominated by value error while its predictions are still uninformative. Training runs under FSDP on 8 GPUs, with rollouts produced by 16 generator shards each driving up to 32 concurrent virtual iOS devices.

\subsection{GRPO}\label{sec:rl-setup}
We initialize the Planning Agent from the 8B vision-language supervised fine-tuning checkpoint described in Section \ref{sec:architecture} and optimize it using Group Relative Policy Optimization (GRPO). No value network is trained; advantages are computed directly from the relative returns within each group of rollouts sampled for the same task. At every training step, we draw 32 tasks and sample $G=4$ trajectories per task, yielding 128 trajectories per step, and each episode is capped at 30 agent steps. Returns are undiscounted ($\gamma=1$, $\lambda=1$). 
Optimization uses AdamW with a constant learning rate of $1\times10^{-6}$, $\beta=(0.9,0.999)$, a weight decay of 0.01, and gradient-norm clipping at 1.0. Rollouts are generated at a temperature of 1.0 and top-$p$ of 1.0, with a maximum of 700 new tokens per step and an 8192-token context. Evaluation rollouts use a temperature of 0.7 and 5 samples per task. Training ran for a single pass over the task set using FSDP (Fully Sharded Data Parallel) for the policy and reference models, supported by the scalable pool of virtual iOS devices.

\textbf{Reward} The reward follows the dense-plus-terminal decomposition outlined in Section \ref{sec:rl}, with the per-step judge returning $+1$ for a success and the terminal judge returning $+10$ for a success and $+5$ for a blocked completion. A trajectory's return is therefore the sum of its per-step rewards plus its terminal reward, and lies within $[0,40]$. Crucially, every reward in this run is non-negative. A step whose plan is judged a failure, whose reasoning is hallucinated, or whose plan cannot be grounded on the current screen receives exactly 0 rather than a penalty. The same is true of a trajectory that exhausts the 30-step budget without completing the task or that keeps acting after the task is already finished. Two consequences follow directly from this design. First, nothing in the objective explicitly discourages unproductive behavior. Because accumulated dense credit is bounded by the step limit of 30, while a terminal success contributes only 10, a long trajectory that collects partial credit can out-earn a short trajectory that actually completes the task. Second, because failure is indistinguishable from inaction at 0 reward, the relative advantage within a GRPO group is determined largely by how much dense credit each rollout happened to accumulate rather than by whether the task was completed. Most importantly, a wrong action is never punished in this experiment. An action that the dense judge rejects, one whose stated reasoning does not match the screen, and one that cannot be grounded at all are all worth the same as no action, which is zero. The policy therefore receives no gradient signal that distinguishes a harmful step from a neutral one, and the only way to change the return is to accumulate more positive credit. Section \ref{sec:rl-optimization} replaces this with a reward in which wrong actions and unsuccessful terminations are explicitly negative.

\subsection{Grounding Experiments}\label{sec:grounding}
To determine the optimal grounding setup for each benchmark, we evaluated three different configurations paired with the SFT baseline planning model: (1) the internal 3B dense model alone, (2) the RL training setup (where the 3B model handles all actions except swipes, which are routed to Gemini-3.5-Flash), and (3) Gemini-3.5-Flash alone. The results are presented in Table \ref{tab:grounding-setup}.

The RL training setup achieved the best performance on UINavBench. This aligns with expectations, as the 3B dense model was trained on real iOS screenshots that match the apps in this benchmark. Furthermore, routing swipes to Gemini-3.5-Flash successfully addresses the 3B model's limitation to direction-only swipes, enabling the location-based swiping required by certain UINavBench tasks.

Gemini-3.5-Flash proved to be the optimal grounding model for iOSWorld. Because iOSWorld consists of synthetic iOS apps, its visual features deviate from the training distribution of the 3B model. Following this logic, since AndroidWorld consists entirely of Android applications, its interfaces deviate even further from the 3B model's training data. As a result, we conclude that Gemini-3.5-Flash is also the best grounding model for AndroidWorld.
 \begin{table}[h]
  \centering
  \begin{tabular}{lccc}
  \toprule
  & 3b & Gemini-3.5-Flash & 3b w/ Gemini-3.5-Flash (swipe only) \\
  \midrule
  UINavBench & 12.7\% & 14.7\% & \textbf{16.7\%} \\
  iOSWorld & 3.0\% & \textbf{4.5\%} & 1.5\% \\
  \bottomrule
  \end{tabular}
  \caption{Performance of the SFT trained Planning agent paired with different grounding models.}
  \label{tab:grounding-setup}
  \end{table}
\subsection{Agent Traces}\label{sec:traces}
\begin{figure}
    \centering
    \includegraphics[width=\linewidth]{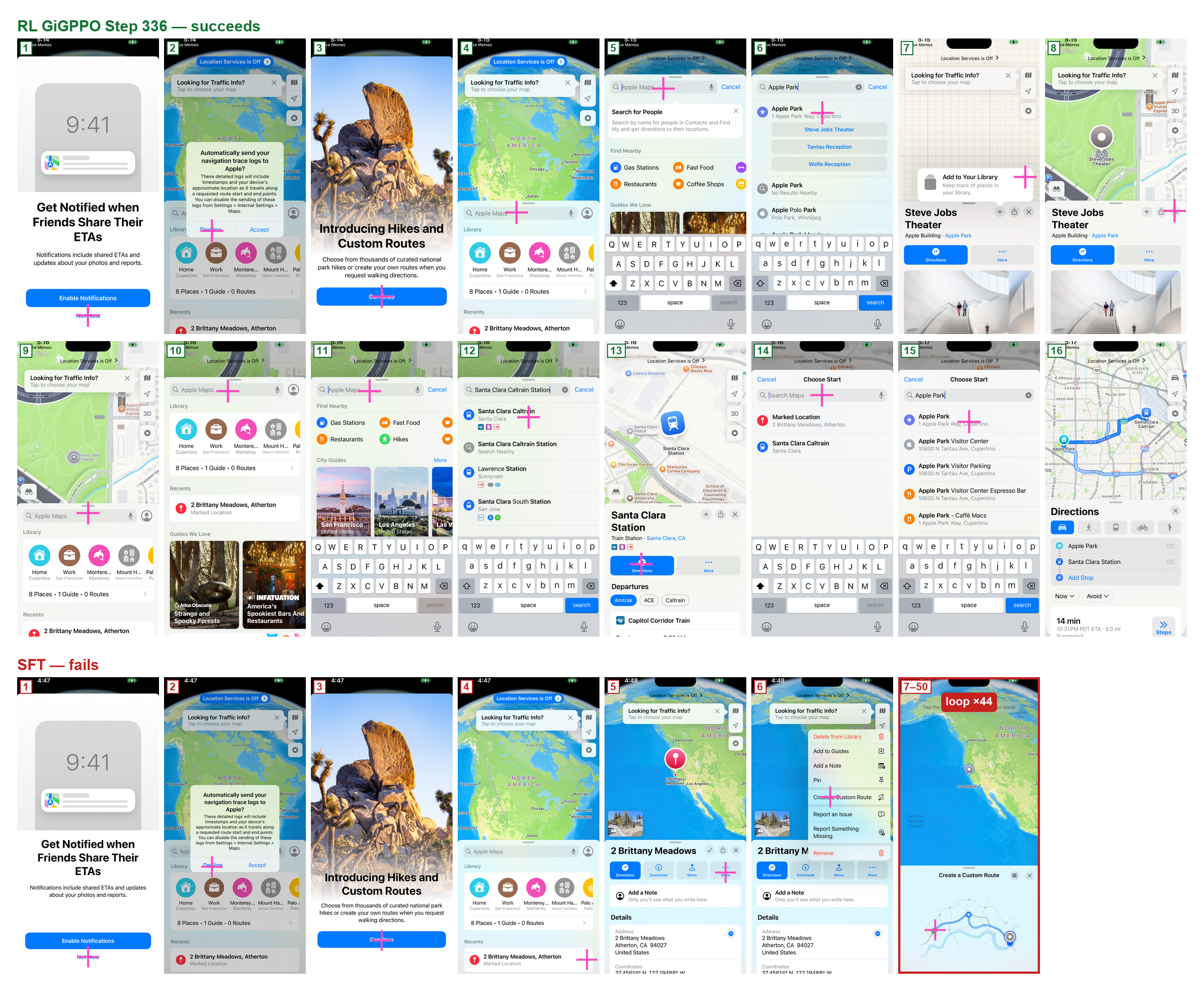}
    \caption{Example traces of the GiGPPO-trained planning agent and the SFT baseline on the UINavBench task: "Find a route from Apple Park to the Santa Clara Caltrain Station." Pink crosshairs indicate the agent's tap and text-entry actions at each step. Because the Apple Maps workflow requires selecting the destination before the starting location, the agents are easily confused by the task prompt's phrasing, which mentions the starting point first. The GiGPPO agent initially sets the destination as Apple Park, but it corrects itself by exiting the directions search and restarting with the Santa Clara Caltrain Station as the destination, demonstrating strong error recovery. In contrast, the SFT agent opens the wrong directions search page and gets stuck repeatedly trying to set the starting point of a custom route until it reaches the maximum step limit.}
    \label{fig:uinavbenchtrace}
\end{figure}

\begin{figure}
    \centering
    \includegraphics[width=\linewidth]{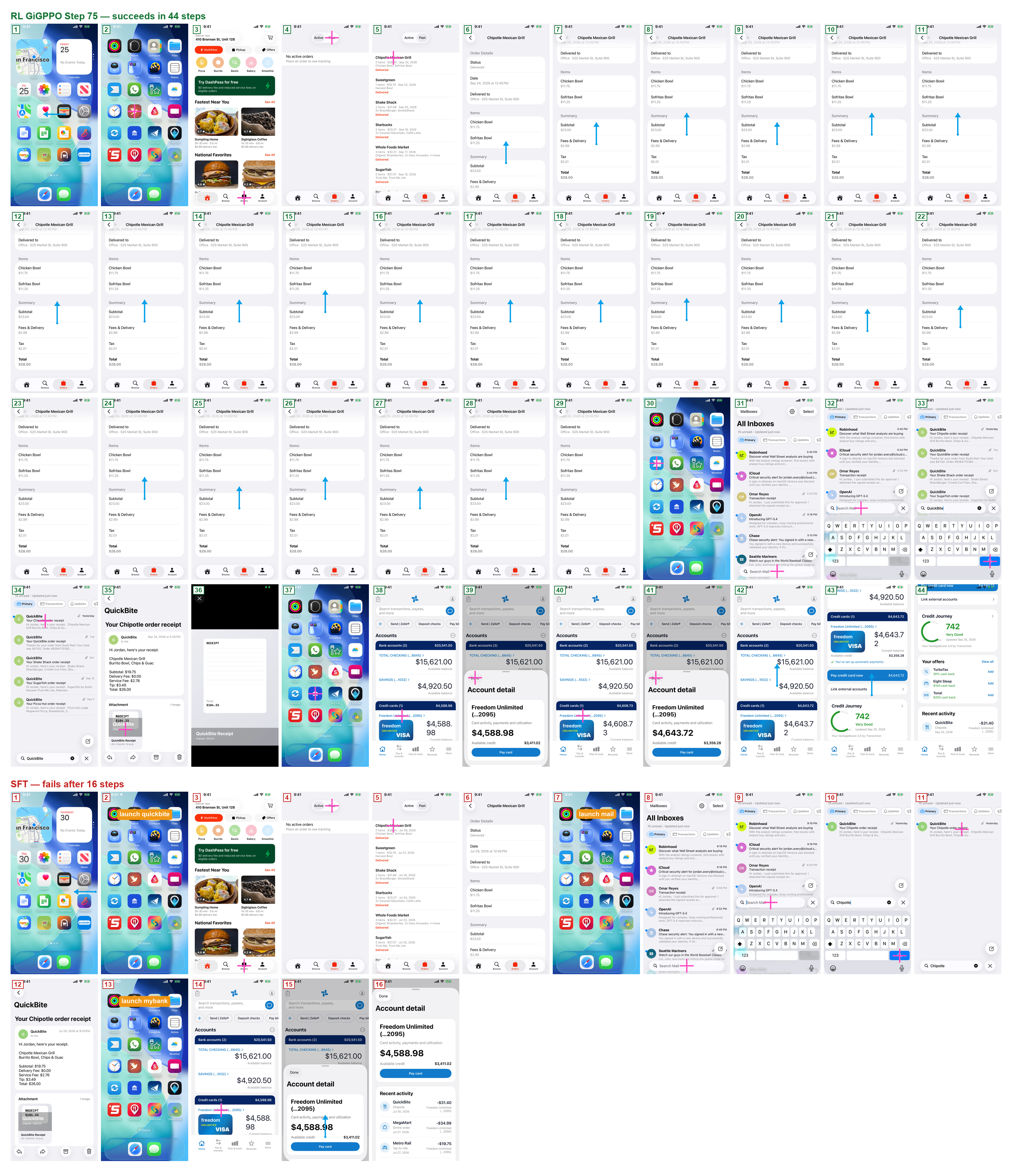}
    \caption{Example traces of the GiGPPO-trained planning agent and the SFT baseline on the complex, multi-app iOSWorld task: ``Find my most recent Chipotle order in QuickBite and note the items and total. Search my Mail inbox for the QuickBite receipt email and note the receipt amount. Check my MyBank credit card transactions for the corresponding charge. Compare the amounts across all three sources (totals may differ due to tips, fees, or processing). What are the order items, QuickBite total, email receipt amount, bank charge, and any discrepancies?'' Pink crosshairs indicate the agent's tap and text-entry actions at each step, while blue arrows indicate swipe actions. Although the SFT agent opens all three apps, it fails to search them for the necessary information. In contrast, the GiGPPO agent successfully navigates through each app and locates all the required details.}
    \label{fig:iosworldtrace}
\end{figure}

\begin{figure}[p]
    \centering
    \includegraphics[width=0.91\linewidth]{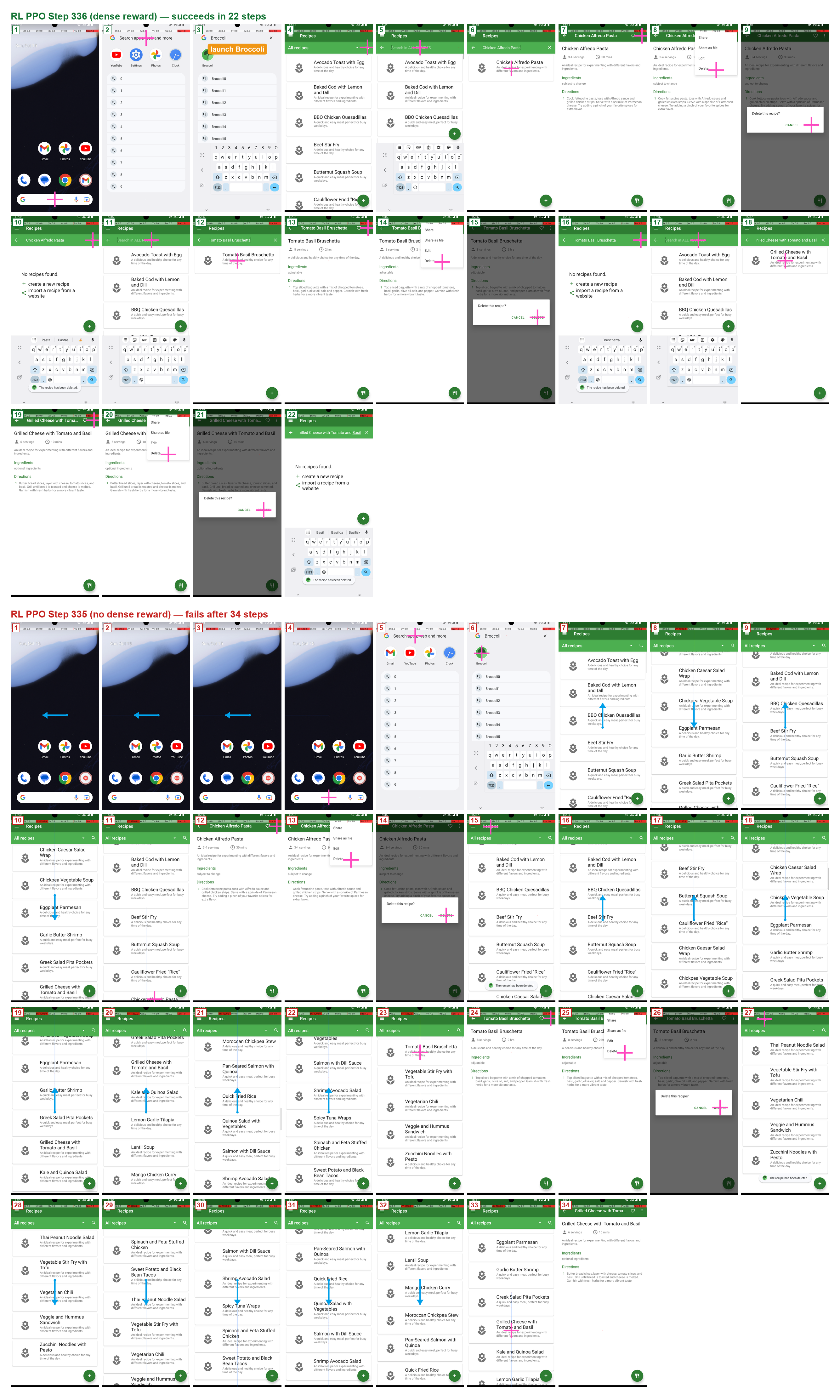}
    \caption{Example traces of the PPO-trained planning agent and the PPO agent trained only on terminal rewards for the AndroidWorld task: "Delete the following recipes from Broccoli app: Chicken Alfredo Pasta, Tomato Basil Bruschetta, Grilled Cheese with Tomato and Basil." Pink crosshairs indicate the agent's tap and text-entry actions at each step, while blue arrows indicate swipe actions. Because the app contains many other recipes, the PPO agent with dense rewards efficiently searches for and deletes each specified recipe. In contrast, the PPO agent without dense rewards manually scrolls through the list to locate each one, wasting numerous steps and failing to complete the task within the maximum step limit.}
    \label{fig:androidworldtrace}
\end{figure}

Figures \ref{fig:uinavbenchtrace}, \ref{fig:iosworldtrace}, and \ref{fig:androidworldtrace} illustrate example traces of the RL-trained agents and the SFT baseline on tasks from UINavBench, iOSWorld, and AndroidWorld. 


\end{document}